# Leveraging Acoustic Cues and Paralinguistic Embeddings to Detect Expression from Voice

*Vikramjit Mitra, Sue Booker, Erik Marchi, David Scott Farrar, Ute Dorothea Peitz, Bridget Cheng, Ermine Teves, Anuj Mehta, Devang Naik*

Apple, Cupertino, CA, USA

vmitra@apple.com

## Abstract

Millions of people reach out to digital assistants such as Siri every day, asking for information, making phone calls, seeking assistance, and much more. The expectation is that such assistants should understand the intent of the user's query. Detecting the intent of a query from a short, isolated utterance is a difficult task. Intent cannot always be obtained from speech-recognized transcriptions. A transcription-driven approach can interpret what has been said but fails to acknowledge how it has been said, and as a consequence, may ignore the expression present in the voice. Our work investigates whether a system can reliably detect vocal expression in queries using acoustic and paralinguistic embedding. Results show that the proposed method offers a relative equal error rate (EER) decrease of 60% compared to a bag-of-word based system, corroborating that expression is significantly represented by vocal attributes, rather than being purely lexical. Addition of emotion embedding helped to reduce the EER by 30% relative to the acoustic embedding, demonstrating the relevance of emotion in expressive voice.

## 1. Introduction

One of the key challenges faced by voice operated assistants, such as Siri, is the interpretation of the intent of the user's query. For example, an intelligent assistant may need to distinguish between a query for information on sports, a request to make a phone call, a command to play music, or many other supported actions. Existing systems exploit spoken language understanding (SLU) to parse user queries into their corresponding intents [1]. Typically, intent detectors are used to assign queries to associated domains for action execution, where such detectors are trained on manually annotated text data [2] [3]. Research on intent detection has primarily focused on leveraging natural language understanding (NLU), where dialog manager [4], intent embeddings [1], joint intent modeling with slot filling [5], and many other approaches have been explored [6] [7]. With the latest advances in machine learning it has become possible to accurately recognize the words in a voice query, resulting in useful intent detection systems.

Speech communication in humans can be broadly split into two layers: (1) a linguistic layer, which conveys the message in the form of words and word-meanings, and (2) a paralinguistic layer, which conveys how those words have been said, including vocal expressiveness. NLU has traditionally focused on (1), but has generally not had access to (2), as such information is embedded in the acoustic signals and is lost to the speech transcription after automatic speech recognition.

The space of possible actions that an intelligent assistant may take is typically divided into functional domains, where each domain provides a subset of actions appropriate to a general area. For example, a domain may exist for making and receiving phone calls, another for reading and sending messages, and another for controlling home automation systems. When a user invokes a digital assistant and speaks a query, the audio is transcribed by the digital assistant's speech recognition system to text. Natural language processing systems parse the text to create an intent, a structure that contains and represents the query's meaning or user's intention. This intent is used in the digital assistant's application logic to surface appropriate dialog, visuals, and additional behavior for users. The work presented here investigates if it is possible to estimate the vocal expression in a query to better understand the intent. Based on the perceived expression, it is possible to improve intent detection performance. For example the intent behind the query to "*find the nearest police station*" may be (a) a general query to seek information, or (b) to make a call, to contact, or seek directions. Vocal expressions can help in better distinguishing between such intents, which may not be obvious from the surface form of the automatic speech transcription.

Vocal expressiveness can be a key signature in detecting the intent of a query. Given two lexically identical queries, lexical features will provide identical representations, but the speech signal retains information on how each query has been expressed and can thus be used to help identify the most appropriate intent. Speech production is a complex cognitive, behavioral, and motor process, where a subtle physiological and cognitive variation due to expression variation can lead to a significant change in the affective state, resulting in noticeable acoustic variations. Such changes affect the speech production mechanism, and can be detected through prosodic, articulatory, and acoustic speech features [8] [9]. A detailed overview of prosodic, articulatory, and acoustic features for detecting speaker state is discussed in [10] [11].

In this work, we investigate whether we can detect the level of expression in a query (in line with human perception). In such a task, we compare how does linguistic (i.e. textual) information perform with respect to the paralinguistic information. Additionally, we seek to answer the following:

- How does perceived emotion (in the form of valence and arousal) correlate with expressiveness?
- Can better acoustic features help to generate better acoustic and emotion embeddings?
- Can articulatory information help in detecting emotional variations in speech?

*Expression* in the context of this paper is defined as vocal expression as perceived by multiple human graders, where inter-grader agreement tended to vary, but not significantly so.

The outline of the paper is as follows: section (2) will present the dataset used in our study, (3) will introduce the speech cues investigated in this paper and the acoustic modeling techniques explored, and (4) we will present the results followed by conclusion in (5).

## 2. Data

We have collected approximately 100 hours of US English speech material and their associated automatically generated speech transcriptions. The data had neither any speaker level information nor any contextual information: every query was independent of every other. Hence the task we explored was: speaker independent and context-free.

The collected speech material was annotated by human graders. After listening to the speech material, the graders answered several questions focusing on the paralinguistic information in speech. Each request was graded by four different graders. In total, 35 graders participated in this task. Graders were trained over a period of 2 to 3 weeks and they were selected based on their performance over a set of mini-grading tests before labeling the data used in this study, to reduce noise in the grades and ensure consistent grading. The grading questions were based on similar tasks previously reported in the literature [12] and internal discussions. The resulting graded dataset contained information on the following attributes:

1. A query's *vocal expression* with respect to the type of intent, i.e., asking for a resource, an accidental trigger, or a prank or other humor attempt. The graders voted if the query expressed the intent clearly, by selecting one of the three possible options: "Yes", "No" and "Not sure".

2. Perceived *primitive emotion* (Arousal and Valence) on a three-level Likert scale.

After grading, the data was filtered to remove cases where all four graders were Not Sure in their decision, which resulted in 70 hours of data that was used in our experiments. The final grade for a query was an average of the individual grades by the graders (where the grades were converted to real integers).

In this work, we also investigated how emotion embedding can help to detect vocal expressions, hence we explored building primitive emotion detection models from speech using the valence and arousal scores from the graded data. The emotion labels from the graded data were averaged across graders and then scaled to have a dynamic range of 1 to 7: (Arousal: 1-very calm, 7-very active; Valence: 1-very negative, 7-very positive).

The graded data was split into 4 groups: (a) pre-training set, 60 hours data; (b) balanced training set (uniform distribution across classes), 30 hours of data; (c) development set (held-out 4 hours of data) and (d) and evaluation set (held-out 3 hours of data).

## 3. Data Analysis and metric

Data grading provided some interesting insights, where the graders agreed more on labeling a query as not-expressive than expressive. Figure 1 shows the distribution of their decisions. The expressive and not-expressive cases are those where two or more graders have agreed strongly toward that decision. When graders labeled primitive emotions, such as perceived valence and arousal levels, the emotion grades tended to have distinct distribution for expressive versus not-expressive queries, as expected, which is shown in figures 2 and 3. Indicating that primitive emotions can assist in detecting the level of expression in a query. The metrics used to measure results on expression detection in this paper are the following (1) equal error rate (EER); (2) f-score; and (3) weighted accuracy (WA) and (4) Unweighted Accuracy (UWA).

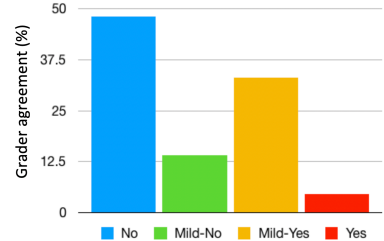

Figure 1: *Distribution of grader agreement on an utterance being expressive: Yes [two or more selected "Yes"]; Mild-Yes [only one selected "Yes" and two or more are "Not Sure"], Mild-No [two or more selected "Not Sure" and the rest selected "No"] and No [two or more selected "No"].*

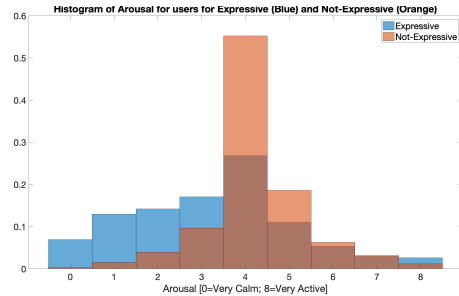

Figure 2: *Distribution of perceived Arousal for expressive versus not-expressive cases.*

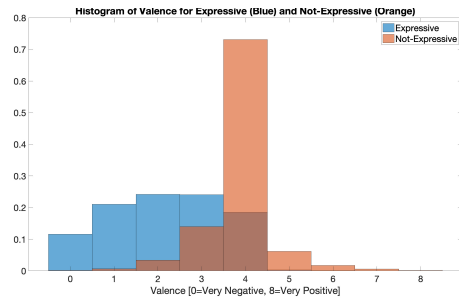

Figure 3: *Distribution of perceived Valence for expressive versus not-expressive cases.*

## 4. Acoustic Features

We investigated several acoustic features to parameterize speech. The baseline feature is the 20 dimensional mel-frequency cepstral coefficients (MFCCs). We explored gammatone cepstral coefficients (GCCs) and modulation features (modulation cepstral coefficients (NMCC) [13]), both of which consisted of 20 cepstral features. In addition, we explored a 3-dimensional pitch, pitch-delta and voicing feature (F0-V). The 3-dimensional F0-V feature was combined with 20 dimen-sional cepstral features (MFCC, GFCC and NMCC) to generate 23 dimensional features (MFCC+F0-V, GFCC+F0-V and NMCC+F0-V).

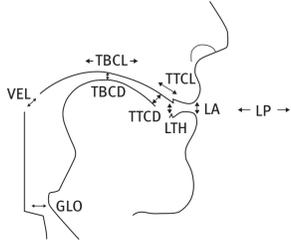

Figure 4: *Mid-sagittal view of the vocal tract constriction variables (TVs) [18].*

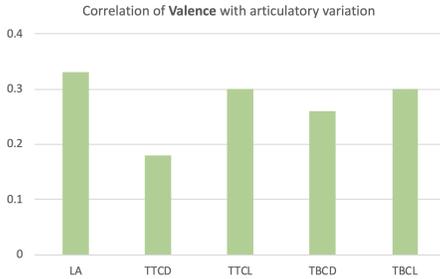

Figure 5: *Correlation of the TVs with valence scores.*

We investigated articulatory features in the form of vocal tract constriction variables (TVs) as detailed in [13]. Detecting valence from speech has been relatively difficult compared to arousal and dominance [14]. Visual and lexical features have been found to be quite useful for improving the performance of valence detection [15]. It has been hypothesized [16] that articulatory information can assist in detecting valence while using speech-only information. This study investigates if valence detection can be improved with the use of articulatory information. The articulatory information (in this case the TVs) in our study defines the degree and location of constriction actions within the human vocal tract as speech is produced. The TVs have eight dimensions: GLO: glottal opening/closing; VEL: velic opening/closing; LP: lip protrusion; LA: lip aperture; TTCL: tongue tip constriction location; TTCD: tongue tip constriction degree, TBCL: tongue body constriction location and TBCD: tongue body constriction degrees [13][17]. A midsagittal view of the vocal tract and its associated TVs are shown in Figure 4 [18]. We analyzed the correlation of the TV trajectory variation with valence and observed positive correlation across the different TV dimensions as shown in Figure 5, justifying the relevance of using TVs for primitive emotion (valence and arousal) detection from speech.

To be able to use TVs in our study, we need to obtain such information from the speech signal, hence we trained an LSTM-based speech-inversion system which takes in spliced (window of 5 frames on both sides of the current frame) 39 dimensional MFCC feature as input and maps that to the 8 TV trajectories. To train the model, we have used the 400 hour articulatory TTS generated data specified in [13].

## 5. Acoustic Model

We used the graded data to train single-layer long-short term memory (LSTM) neural network based acoustic models, with

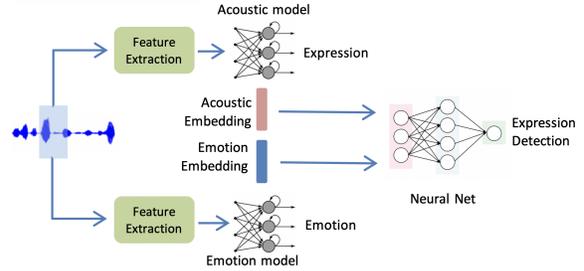

Figure 6: *Embedding fusion for expression detection.*

128 neurons in the recurrent and the embedding layers. The models were tuned using a held-out dev set. The models were trained using cross-entropy loss, with a mini-batch size of 200, the Adam optimizer, and a momentum of 0.9. Given the imbalance of classes (as evident in Figure 1), the training data was grouped into two: (1) pre-training data, consisting of all the graded data and (2) fine-tuning data, a balanced subgroup of the graded data, where each class was equally likely. The learning rate used during pre-training and fine-tuning was 0.0001 and 0.01 respectively. In addition to using LSTM models for directly detecting expression from speech, we also investigated training emotion embedding networks for valence and arousal. The recurrent and embedding layers of the LSTM network consisted of 64 neurons. The models were trained using mean-squared error loss, with a mini-batch size of 300, using Adam optimizer, and a momentum of 0.9, with a learning rate of 0.01.

For all the model training steps, early stopping was allowed based on cross-validation error increase. If the cross-validation step increased for five consecutive steps, the training step backed-off to the previous best model and restarted training with a 10% reduced learning rate. If such back-off strategy failed, then training stopped, returning the most recent best model, based on training and cross-validation error. Finally, a single hidden layer feed-forward neural network with 128 neurons (shown in Figure 6) was used to take the combination of the acoustic embedding and emotion embedding as input and generate the expression scores. This network was trained with the same training and cross-validation list as was used to train the LSTM models.

## 6. Results

We investigated text-based models for the given task, where bag-of-words (BoW) features were used to train a multi-layered neural network (NN). We also used a random model that generated random outputs and the resulting scores are shown in Table 1. Both the number of hidden layers and number of neurons in each layer were optimized given a held-out validation set. The BoW feature transforms were learned from the speech transcriptions of the 60-hour pre-training data, and the neural net model was trained using BoW features obtained from the 30-hour balanced data. Additionally, an MFCC feature based LSTM model was trained as a baseline acoustic model and the results from the text-only and audio-only models are shown in Table 1. The baseline acoustic model (MFCC-LSTM) performed better than the text-based model (BoW-NN), by demonstrating a significantly lower EER and higher WA, with a confidence interval of 100%. Table 1 clearly shows the weakness of the text-based system which performs almost at chance level, where WA was close to 50%. This is because two queries may have the exact same text, but their expression can be quite different, indicating that expressive cues are present at the acoustic level but not at the lexical level. While the general intent of a query may reside in its surface form, its expression is embedded in its acoustic signal as paralinguistic information. Next, we investigated the use of different acoustic features as input to the LSTM acoustic model and the results obtained from those experiments are shown in Table 2. Table 2 shows an interesting trend, where EER is found to improve when the pitch features (F0-V) are combined with the cepstral features (MFCC, GCC, and NMCC) compared to the cepstral features alone. Similar trend is seen in F-score as well, indicating that pitch and voicing contour is useful in capturing expressive cues in speech.

Table 1: *Equal error rate (EER) and weighted accuracy (WA) from a random model, text-based and audio-based models*

| Feature | Model | EER | WA(%) |
|---|---|---|---|
| Random | - | 50.74 | 49.69 |
| BoW | DNN | 47.46 | 52.79 |
| MFCC | LSTM | 29.05 | 65.34 |

Table 2: *EER, WA and, F-score from different acoustic features as input to the LSTM acoustic model.*

| Feature | EER (%) | WA (%) | F-score |
|---|---|---|---|
| MFCC | 29.05 | 65.34 | 0.64 |
| GCC | 30.15 | 68.82 | 0.66 |
| NMCC | 29.18 | 67.98 | 0.65 |
| MFCC+F0-V | 28.23 | **73.51** | 0.66 |
| GCC+F0-V | 27.33 | **73.51** | 0.67 |
| NMCC+F0-V | **27.11** | 72.03 | **0.68** |

As noted in Section 4 (Figure 2), emotion related information can help discern vocal expression. For example, expressive queries seem to have a different valence and arousal range than the non-expressive ones. To investigate whether the emotional content of a query can help in the given task, we trained an LSTM model for predicting valence and arousal from speech. The emotion LSTM model takes in the same set of features as the expression acoustic models. The results from the emotion LSTM models are shown in Table 3, where concordance correlation coefficient (CCC) is used as the evaluation metric.

We extracted the embeddings from the MFCC+F0-V and NMCC+F0-V expression acoustic models and used them as input to a single hidden layer neural network (NN) with 256 neurons, where the targets were the expression scores. Table 2 shows the EER from the acoustic embedding trained NN models. In addition, we combined the acoustic embeddings (AE) from models trained with NMCC+F0-V and MFCC+F0-V features, and the emotion embeddings (EE) from the model trained with MFCC+F0-V+TV feature, and the results are shown in Table 4. Figure 7 shows the ROC curve from the random, BoW, AE and embedding fusion (AE+EE and AE2+EE, shown in 3rd and 4th rows of Table 4, respectively) systems.

Table 4 shows the following: (1) better acoustic features generated better embedding (NMCC), which in turn helped to improve performance as compared to other acoustic features (such as MFCC). (2) The addition of emotion embedding helped to reduce the EER by 26% relative with respect to the best performing single feature system (NMCC+F0-V). In addition, a relative reduction of 34% in EER is achieved compared to the MFCC baseline and an overall 59% relative reduction in EER is observed compared to the text-only system.

Table 3: *Performance of the various Emotion LSTM models w.r.t concordance correlation coefficient (CCC).*

| Feature | CCC | |
|---|---|---|
|  | Valence | Arousal |
| MFCC | 0.27 | 0.64 |
| GCC | 0.28 | 0.67 |
| NMCC | 0.26 | **0.68** |
| MFCC+F0-V | 0.37 | 0.66 |
| MFCC+F0-V+TV | **0.40** | 0.66 |

Table 4: *Performance from using Acoustic Embedding (AE) and Emotion Embedding (EE) in single layered neural net*

| Embeddings | EER |
|---|---|
| AE(MFCC+F0-V) | 28.31 |
| AE(NMCC+F0-V) | 27.26 |
| AE(NMCC+F0-V) +EE(MFCC+F0-V+TV) | 20.01 |
| AE2(NMCC+F0-V, MFCC+F0-V) +EE(MFCC+F0-V+TV) | **18.84** |

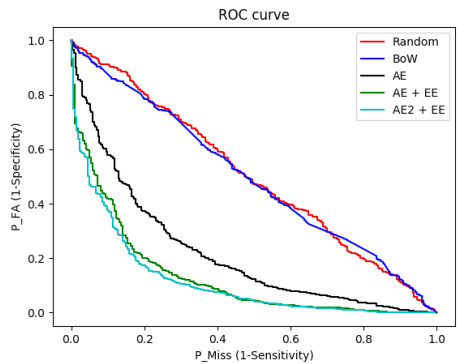

Figure 7: *ROC curve from the random, BoW, AE and EE systems.*

## 7. Conclusions

In this work, we investigated how acoustic and emotion cues can be used to detect vocal expression in speech. We observed that (a) primitive emotion can help in determining vocal expression, (b) articulatory information can help in improving the valence detection, and (c) robust acoustic features can help in generating better embedding. We have presented evidence that vocal expression is a useful acoustic attribute that could complement lexical information, where acoustic based systems demonstrated a significant error rate reduction compared to text-only systems. In the future, we plan to investigate attention modeling to focus on speech only regions which can help to improve the performance of the given task.

## 8. Acknowledgements

The authors would like to thank Russ Webb, Sachin Kajarekar and Alex Acero for their valuable comments and suggestions to improve the contents of this paper.